\documentclass[letterpaper, 10 pt, conference]{ieeeconf}  

\IEEEoverridecommandlockouts
\usepackage{url}
\usepackage{balance}
\usepackage{amsmath}
\usepackage{amssymb,bm}
\usepackage{amsfonts}
\usepackage{enumerate}
\usepackage{tabulary}
\usepackage{multirow}
\usepackage{cite}
\usepackage{lipsum}  
\usepackage{graphicx}
\usepackage{epstopdf}
\usepackage[misc]{ifsym}
\usepackage{color}
\usepackage{xcolor}
\usepackage{xxcolor}
\usepackage{pgf}
\usepackage{pgfplots}
\usepackage{tikz}
\usepackage{float}
\usepackage{lipsum}
\usepackage{booktabs}
\usepackage{verbatim}
\usepackage{hyperref}
\usepackage{makecell}
\usepackage{xfrac}

\usepgfplotslibrary{groupplots,fillbetween}
\usepackage{pgfplotstable}

\pgfplotsset{compat=1.18}

\usepackage{ifthen}
\usepackage{forloop}
\usepackage{listings}
\usepackage[lined,boxed,commentsnumbered,linesnumbered]{algorithm2e}

\definecolor{codegreen}{rgb}{0,0.6,0}
\definecolor{codepurple}{rgb}{0.58,0,0.82}
\definecolor{backcolour}{rgb}{0.95,0.95,0.92}
\lstdefinestyle{buzz}{
    backgroundcolor=\color{black!5},   
    commentstyle=\color{codegreen},
    keywordstyle=\color{blue},
    numberstyle=\tiny\color{black!30},
    stringstyle=\color{codepurple},
    basicstyle=\footnotesize\ttfamily,
    breakatwhitespace=false,         
    breaklines=true,                 
    captionpos=b,                    
    keepspaces=true,                 
    numbers=left,                    
    numbersep=5pt,                  
    showspaces=false,                
    showstringspaces=false,
    showtabs=false,                  
    tabsize=2,
}
\newcommand{\transpose}{\intercal}

\SetAlCapSkip{0.5em}

\renewcommand\vec{}

\title{\LARGE \bf
A Remote Sim2real Aerial Competition: Fostering\\
Reproducibility and Solutions' Diversity in Robotics Challenges
}

\author{
Spencer Teetaert$^1$,
Wenda Zhao$^1$,
Niu Xinyuan$^2$, Hashir Zahir$^2$, Huiyu Leong$^2$,\\
Michel Hidalgo$^3$, Gerardo Puga$^3$, Tomas Lorente$^3$, Nahuel Espinosa$^3$, John Alejandro Duarte Carrasco$^3$,\\
Kaizheng Zhang$^4$, Jian Di$^4$, Tao Jin$^4$, Xiaohan Li$^4$, Yijia Zhou$^4$, Xiuhua Liang$^4$, Chenxu Zhang$^4$,\\
Antonio Loquercio$^5$,
Siqi Zhou$^{1,6}$,
Lukas Brunke$^{1,6}$,
Melissa Greeff$^1$,\\
Wolfgang H\"onig$^7$,
Jacopo Panerati$^1$,
and Angela P. Schoellig$^{1,6}$
\thanks{
$^1$\href{http://www.dynsyslab.org}{Learning Systems \& Robotics Lab (LSY)}, Institute for Aerospace Studies, University of Toronto and the Vector Institute for Artificial Intelligence;
$^2$Team H$^2$, Singapore; 
$^3$Team Ekumen, Argentina; 
$^4$University of Science and Technology of China; 
$^5$University of California, Berkeley;
$^6$Technische Universit\"at M\"unchen;
$^7$Technische Universit\"at Berlin
}
}

\begin{document}
\maketitle
\thispagestyle{empty}
\pagestyle{empty}

\begin{abstract}
Shared benchmark problems have historically been a fundamental driver of progress for scientific communities.
In the context of academic conferences, competitions offer the opportunity to researchers with different origins, backgrounds, and levels of seniority to quantitatively compare their ideas. 
In robotics, a hot and challenging topic is  \emph{sim2real}---porting approaches that work well in simulation to real robot hardware.
In our case, creating a hybrid competition with both simulation and real robot components was also dictated by the uncertainties around travel and logistics in the post-COVID-19 world.
Hence, this article motivates and describes an aerial sim2real robot competition that ran during the 2022 IEEE/RSJ International Conference on Intelligent Robots and Systems, from the specification of the competition task, to the details of the software infrastructure supporting simulation and real-life experiments, to the approaches of the top-placed teams and the lessons learned by participants and organizers.
\end{abstract}

\section{Introduction}
\label{sec:intro}

Advances in robotics promise improved functionality, efficiency, and quality with an impact on many aspects of our daily lives. Example applications include autonomous driving, drone delivery, and service robots. However, the decision-making of such systems often faces multiple sources of uncertainty (e.g., incomplete sensory information, uncertainties in the environment, interaction with other agents, etc.). Deploying an embodied autonomous learning system in real-world~\cite{DSL2021,Hewing2020,mitchell2020} and possibly commercial applications requires both \emph{(i)} safety guarantees that the system acts reliably in the presence of the various sources of uncertainties and \emph{(ii)} efficient deployment of the decision-making algorithm to the physical world (for performance and cost-effectiveness). As highlighted in the “Roadmap for US Robotics”, learning and adaptation are essential for next-generation robotics applications, and guaranteeing safety is an integral part of this.

With this competition, which ran virtually during the 2022 IEEE/RSJ International Conference on Intelligent Robots and Systems (IROS), our goal is to bring together researchers from different communities to \emph{(i)} solicit novel and data-efficient robot learning algorithms, \emph{(ii)} establish a common forum to compare control and reinforcement learning approaches for safe robot decision-making~\cite{DSL2021}, and \emph{(iii)} identify the shortcomings or bottlenecks of the state-of-the-art algorithms with respect to real-world deployment.

\begin{figure}
    \centering
    \begin{tikzpicture}
    \begin{groupplot}[
    group style = {group size = 1 by 1, horizontal sep=1.5cm, vertical sep=1.5cm,},
    width = 1.0\columnwidth, 
    height = 6.50cm,
    xmin=0, 
    xmax=100, 
    ymin=0, 
    ymax=100,
    grid=none, axis line style={draw=none}, 
    tick style={draw=none}, 
    clip=false,
    xticklabel=\empty, 
    yticklabel=\empty,
    axis on top=true,
    clip marker paths=true,
    ]

    \nextgroupplot[]
        \node[opacity=1.0] at (axis cs:50,50) () {\includegraphics[draft=False,width=.99\columnwidth]{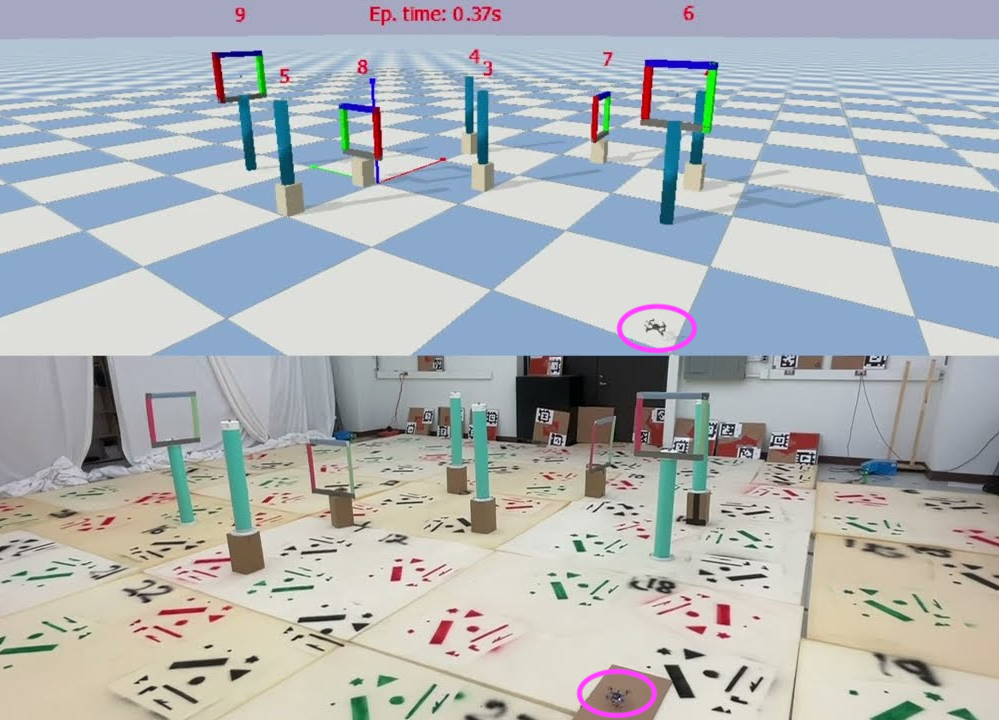}};

    \end{groupplot}
\end{tikzpicture}
\vspace{-1em}     \caption{Examples of the competition's obstacle course in simulation (PyBullet, top) and in the real world (University of Toronto, bottom).}
    \label{fig:setup}
\end{figure}

In this article, we report on three major contributions made by the organizers and participants of the competition:
\begin{itemize}
    \item The competition design, intended to stress and evaluate the controllers proposed by the participant teams along multiple axes---safety, robustness, adaptiveness, and learning---using multiple levels of increasing complexity.
    \item The implementation details of the sim2real infrastructure used to let participants enter the competition remotely while also having their solution evaluated in real flying robotic hardware. Importantly, this helped to foster diversity in the origin, background, and seniority of the competition's participants.
    \item The theoretical and implementation details---as well as the quantitative evaluation of the sim2real transfer performance---of the solutions of the top 3 participant teams---University of Science and Technology of China, Ekumen, and H$^2$.
\end{itemize}

The rest of the article is organized as follows.
Section~\ref{sec:related} highlights the importance and value of robotic competition for the research community.
Section~\ref{sec:comp} outlines the competition task and the implementation of the simulation and sim2real frameworks used to develop and evaluate the participants' solutions.
Sections~\ref{sec:ustc}, \ref{sec:ekuflie}, and \ref{sec:h2} are contributed by the three teams that reached the competition podium and they detail the methods and insights behind each solution.
Finally, Sections~\ref{sec:sim2real-results}
and~\ref{sec:lessons} summarizes the competition's results and the lessons learned by participants and organizers.

 \begin{table*}
	\centering
	\caption{Simulated and Real-world Evaluation Scenarios of the Competition}
	\begin{tabular}{cccccl}
    \toprule
Evaluation Scenario & Constraints & Rand. Inertial Properties & Randomized Obstacles, Gates & Rand. Between Episodes & Notes \\
\cmidrule(lr){1-6}
\href{https://github.com/utiasDSL/safe-control-gym/blob/beta-iros-competition/competition/level0.yaml}{level 0} & \textbf{Yes} & \emph{No} & \emph{No} & \emph{No} & Perfect knowledge \\
\href{https://github.com/utiasDSL/safe-control-gym/blob/beta-iros-competition/competition/level1.yaml}{level 1} & \textbf{Yes} & \textbf{Yes} & \emph{No} & \emph{No} & Adaptive \\
\href{https://github.com/utiasDSL/safe-control-gym/blob/beta-iros-competition/competition/level2.yaml}{level 2} & \textbf{Yes} & \textbf{Yes} & \textbf{Yes} & \emph{No} & Learning, re-planning \\
\href{https://github.com/utiasDSL/safe-control-gym/blob/beta-iros-competition/competition/level3.yaml}{level 3} & \textbf{Yes} & \textbf{Yes} & \textbf{Yes} & \textbf{Yes} & Robustness \\
\cmidrule(lr){1-6}
\href{https://www.youtube.com/watch?v=PwphA_jsNKw}{sim2real} & \textbf{Yes} & Real-life hardware &  \textbf{Yes}, injected & \emph{No} & Sim2real transfer \\
    \bottomrule
\end{tabular}
\newline \hfill \newline
{\footnotesize
Note: ``Rand. Between Episodes'' states whether randomized properties and positions vary or are kept constant across episodes.
}
\vspace{-2em}
 	\label{tbl:scenarios}
\end{table*}

\section{Robot Competitions}
\label{sec:related}

Robotics competitions are essential for two main reasons: they are catalysts for progress in robotics research and have great educational value.
First, robotics competitions provide a standardized platform for researchers to test and showcase their latest innovations. To maximize research progress, competitions use three main tools: \emph{(i)} standard metrics, \emph{(ii)} sequestered test data, to avoid solutions overfitting to the test set, and \emph{(iii)} public leaderboards and reports (such as this one), to understand what are the key ideas behind the most successful entries. These tools enable objective comparisons between approaches, speeding up progress.

The second key value of competitions is education. They provide an opportunity for students to learn about robotics and apply their knowledge hands-on while promoting collaboration and teamwork. On top of their educational value, competitions represent a medium for students and organizers to network beyond universities, creating a community.

Given these benefits, it comes as no surprise that the aerial robotics community has organized and continues to develop multiple competitions. These competitions can greatly differ in scale. Some are very large-scale, attracting both private and government investment. Examples include the DARPA Fast Lightweight Autonomy (FLA)~\cite{darpa_FLA} program or the MBZIRC aerial robotic league~\cite{mbzirc}. Such large-scale efforts are very important to foster robotics progress, but they have a smaller impact on education.

Conversely, smaller-scale competitions significantly impact education, since they provide opportunities for young students to design and build full-stack robotics systems. Examples are the DodgeDrone series~\cite{song2022dodgedrone} and the sim2real IROS competition presented herein (Figure~\ref{fig:setup}). 

\section{IROS 2022 Sim2real Competition}
\label{sec:comp}

\subsection{Task and Evaluation Scenarios}
\label{sec:task}

The competition task is to design (in simulation) a controller/planner that enables a nano-quadrotor (specifically, the Bitcraze Crazyflie 2.x) to safely fly (in the real world) through a set of gates and reach a predefined target despite uncertainties in the robot dynamics (i.e., mass and inertia), in the robot actuation (i.e., input disturbances), and the environment (i.e., wind and position of the gates). 

The solutions are evaluated in terms of both their safety (i.e., producing no collisions with gates or obstacles) and performance (i.e., time to reach and stabilize at the target position, having navigated all the gates in the given order). The competition task was designed to encourage the exploration of both control and reinforcement learning approaches (e.g., robust, adaptive, predictive, learning-based and optimal control, and model-based/model-free reinforcement learning). 

The controller/planner of each solution has access to the position and attitude measurements (provided by the simulation engine or a motion capture system in real-life experiments) and the noisy pose of the closest next gate. To mimic a limited-range perception system, the exact gate pose is only revealed to the controller once the quadrotor is within less than half a meter of it. The controller can send position, velocity, acceleration, and heading references to an onboard position controller.

All the proposed solutions were evaluated in 5 scenarios with different challenges, culminating in real-world evaluations (see Table~\ref{tbl:scenarios}). Constraints included both state (do not leave the boundaries of the flight arena) and input (feasible action) constraints.

Each solution was required to re-implement 2 methods (the initialization and the action selection of the \texttt{\small Controller} class) and, optionally, could re-implement 2 more methods allowing learning between control steps (\texttt{\small interStepLearn()}) and between training episodes (\texttt{\small interEpisodeLearn()}).

\subsubsection{Scoring System}
\label{sec:score}

For all levels in Table~\ref{tbl:scenarios} (0-3, sim2real), solutions were evaluated---on the last episode of each level---by:
\begin{itemize}
    \item \emph{safety}, i.e., the ability to avoid \emph{(i)} all collisions with gates and obstacles and \emph{(ii)} all constraint violations (only episodes with 0 collisions and violations are considered as successful task completions);
    \item and \emph{performance}, i.e., minimizing the task time (in seconds) required to complete the task (flying through all the gates, reaching, and stabilizing at the goal position).
\end{itemize}

For all levels in Table~\ref{tbl:scenarios} (0-3, sim2real), solutions that accomplish the task (as described in the previous paragraph) were also evaluated---across all episodes in each level---by:
\begin{itemize}
    \item  \emph{data \& compute efficiency}, i.e., minimizing the simulation-clock/flight time (in seconds) plus the overall wall-clock learning time (in seconds) used by methods \texttt{\small interStepLearn()} and \texttt{\small interEpisodeLearn()}.
\end{itemize}

For all levels, the top 3 solutions ranked by \emph{safety} and \emph{performance} and the top 3 solutions ranked by \emph{data \& compute efficiency} were given 20, 10, and 5 points respectively. The sum of these points determined the final classification (see Table~\ref{table:rank} and this \href{https://www.youtube.com/watch?v=C6PZYJ5R1MI}{\bfseries{video of the top 3 solutions}}.

\subsection{Implementation Details}
\label{sec:impl}

One of the main contributions of our work is to break away from both purely simulated and in-person-only competitions by creating a framework for the remote and asynchronous development of novel robot controllers.

\subsubsection{Sim2real}
\label{sec:sim2real}

We developed a novel software framework to minimize the differences between simulation and real-world application, enabling controllers designed in simulation to be tested on flight hardware without the need for fine-tuning. Our \emph{sim2real} framework consists of three parts: 
\begin{itemize}
    \item \texttt{\small pycffirmware}\protect\footnote{https://github.com/utiasDSL/pycffirmware}: an extension of Bitcraze's Crazyflie firmware\footnote{https://github.com/bitcraze/crazyflie-firmware} Python bindings;
    \item a firmware wrapper module to interface between \texttt{\small pycffirmware} and \texttt{\small safe-control-gym}\protect\footnote{https://github.com/utiasDSL/safe-control-gym};
    \item and a module to execute controllers on the flight hardware using \emph{Crazyswarm}\protect\footnote{https://github.com/USC-ACTLab/crazyswarm}.
\end{itemize}

The firmware wrapper was designed to use the Crazyswarm API so that controllers could be copy-and-pasted \emph{as is} between simulation and flight hardware. The module wraps the Crazyflie firmware to ensure onboard changes to command signals are captured in \texttt{\small safe-control-gym}~\cite{safecontrolgym}. 

Our goal was to let the competition's submissions be developed using the sim2real pipeline in simulation; and then, to be able to test the different solutions in both simulation and on flight hardware. 

\begin{figure}
    \centering
    \includegraphics[]{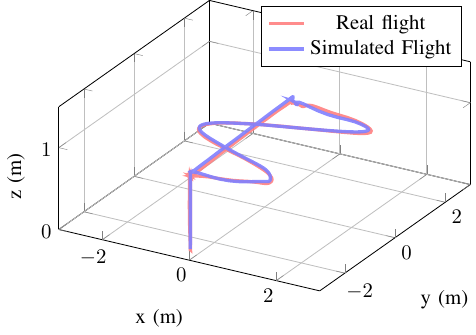}

     \caption{Example of sim2real performance comparing simulated (blue) and real-world (red) trajectories.}
    \label{fig:sim2real}
\end{figure}

To validate the sim2real pipeline, we developed 7 test controllers and trajectories in simulation and flew each one of them 30 times on real Crazyflies (e.g., Figure~\ref{fig:sim2real}). The error between the simulated flight path and real-world flight path varied by less than 5cm on average, with total flight path lengths varying from less than 2m to greater than 10m.  

\subsubsection{\texttt{\small pycffirmware}}
\label{sec:pycf}

We developed an extension to the Crazyflie Python bindings to facilitate full wrapping of the functional elements of the Crazyflie's firmware, that is, modules in the firmware that limit execution rates or change a reference signal's value. For this competition, we required teams to use the firmware's Mellinger controller. We note that \texttt{\small pycffirmware} was developed in parallel to extensions of Bitcraze's official Python binding that are used in the Crazyflie Webot simulation wrapper\footnote{https://github.com/bitcraze/crazyflie-simulation} and Crazyswarm2\footnote{https://github.com/IMRCLab/crazyswarm2}, resulting in a functional overlap between the two approaches.

\subsubsection{\texttt{\small safe-control-gym}}
\label{sec:gym}

We developed a firmware wrapper module for \texttt{\small safe-control-gym}~\cite{safecontrolgym} that interfaces with the \texttt{\small pycffirmware} bindings. This provides an interface between the firmware variables and the simulation physics engine. A quadratic mapping is used to convert motor PWM commands from the firmware to thrust magnitudes used by \texttt{\small safe-control-gym}~\cite{panerati2021learning}. The competition included timing hints on the execution of each new controller but did not enforce a hard cutoff, allowing the development of computationally expensive solutions. These would perform above average in simulation but also incur a poorer sim2real transfer.

\subsubsection{Crazyswarm}
\label{sec:crazyswarm}

To ensure compatibility between the sim2real implementation in \texttt{\small safe-control-gym} and the real-world implementation we limited the use of the Crazyswarm~\cite{crazyswarm} Python API to functions that we validated in \texttt{\small safe-control-gym}. This includes \texttt{\small goTo()}, \texttt{\small cmdFullState()}, \texttt{\small takeoff()}, \texttt{\small land()}, and \texttt{\small stop()} from the \texttt{\small Crazyflie} class. Commands were sent from a ground station to the Crazyflie robot at a rate of 30Hz. We used a Vicon motion tracking system to measure the locations of each obstacle at the start of each trial and provide accurate information to the controller of each proposed solution.

The following three sections detail the work of the teams whose solutions reached the podium of the competition. 
\section{Team USTC's Solution}
\label{sec:ustc}

Team USTC is from the Department of Automation in the School of Information Science and Technology of USTC, under the guidance of Professor Haibo Ji and Professor Xinghu Wang, and mainly studies flight control and trajectory optimization. Multiple original techniques are used to solve the specific problems encountered in the competition scenario, which is introduced in Section~\ref{sec:comp}.

In the planning phase, team USTC uses a novel sampling-based planning scheme to find a collision-free path and then generate a feasible trajectory using the minimum snap method with time allocation that passes through the set of waypoints \cite{Richter2016}. In the controlling phase, \emph{(i)} a learning-based compensation approach is used to reduce the control errors caused by quadrotor mass uncertainty, and \emph{(ii)} an event-triggered trajectory adjusting method is used to modify the planned reference trajectory according to the gates’ actual positions.

This section briefly summarizes the solution's main ideas, implementation details, and experiment results.

\subsection{Methods}
\label{sec:ustc-methods}
\begin{figure}
    \centering
    \includegraphics[draft=False,width=.99\columnwidth,
    height=5.5cm
    ]{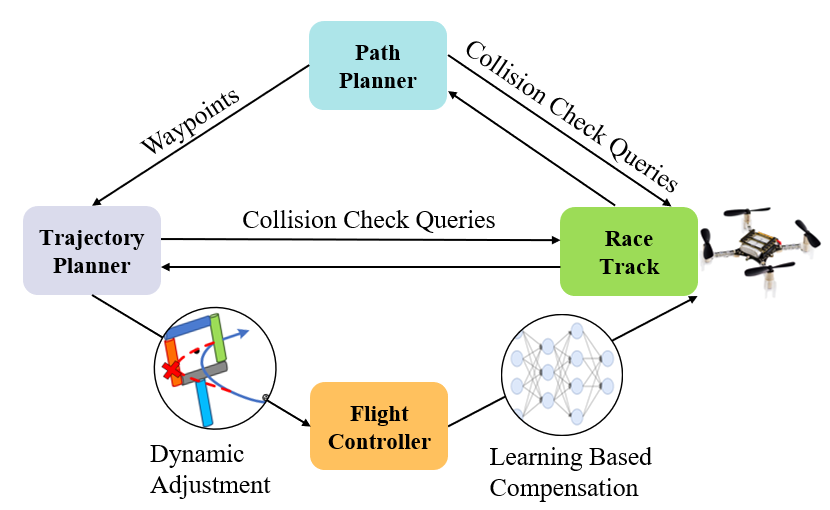}
    \caption{System diagram of team USTC.}
    \label{fig:ustc}
\end{figure}

Figure~\ref{fig:ustc} presents a system diagram of the proposed approach.
The path planner module generates a global collision-free path. The trajectory planner connects the path points with time-based polynomials to form a smooth trajectory. The learning-based compensation module enables the controller to estimate the error in the quadrotor mass during the flight. The dynamic adjustment module performs the real-time reference trajectory correction to guide the quadrotor through the real center of the next gate, which is obtained when the quadrotor is sufficiently close to it.

\begin{figure}
    \centering
    \includegraphics[scale=0.15]{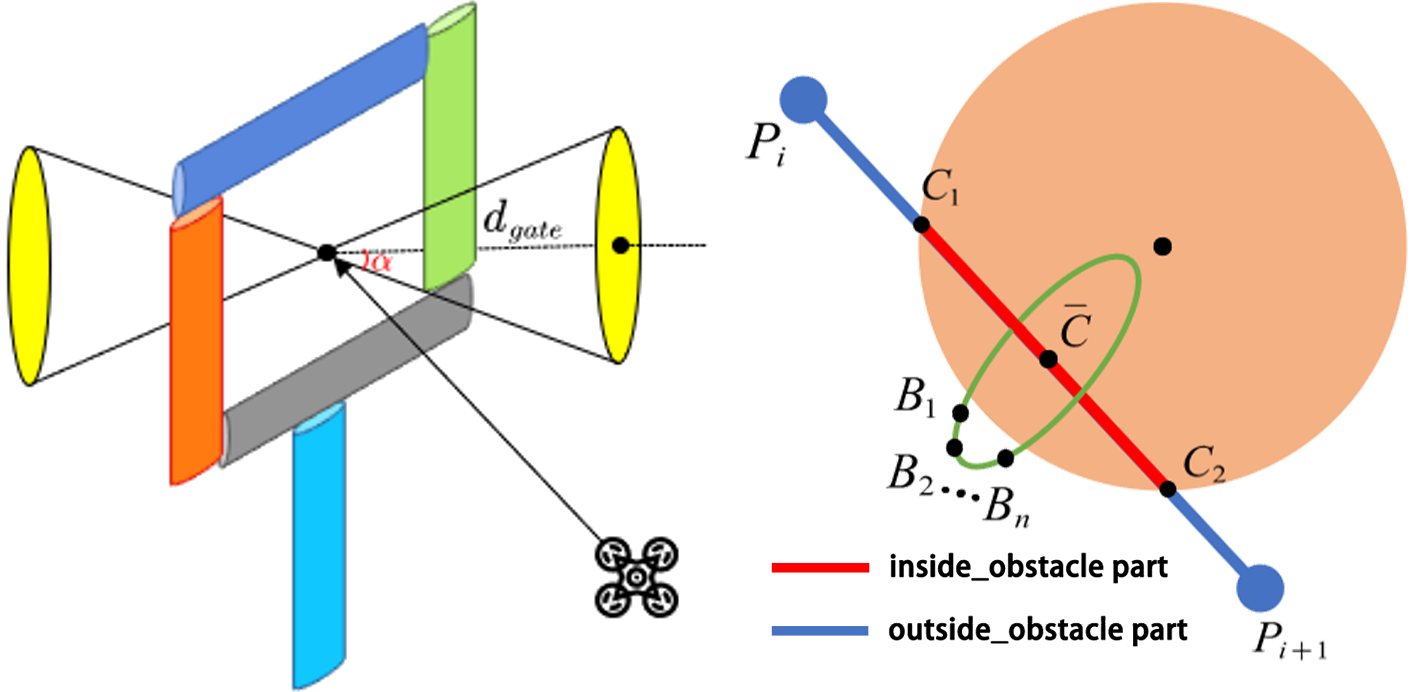}
    \caption{An illustration of the sampling-based path points construction method. The diagram on the left shows the sampling point in front of the gate when the angle between the direction of the path and the direction of the gate exceeds a threshold. The diagram on the right illustrates the sampling process when the segment between two adjacent path points intersects with an obstacle.}
    \label{fig:ustc-inf}
\end{figure}

\subsubsection{Path Planner}
\label{sec:path}

The novel sampling-based path generation method is the core of the path planner module. Here, a path $\mathcal{P}$ is denoted as a discrete sequence of positional points in the 3D space, $\mathcal{P}=\left\{ P_1,P_2,\dots, P_k \right\}$. During the path generation step, we first concatenate the coordinates of the center point of each gate to form an initial path. Then, we perform a feasibility analysis along the initial path. There are two possible kinds of infeasibility. One is when the angle between the direction of the path and the direction of the next gate exceeds a threshold $\alpha$, and the other is when the segment between two adjacent path points intersects with an obstacle. When the first kind occurs, a new waypoint is added in front of the corresponding gate at a distance of $d_{\mathrm{gate}}$ meters. When the second kind occurs, all intersection points that split the segment into inside-obstacle parts and outside-obstacle parts are recorded. For every inside-obstacle part in Figure~\ref{fig:ustc-inf}, we calculate its midpoint and perform sampling along a circle that is centered at the midpoint and perpendicular to the segment.
We gradually increase the sampling radius until a collision-free point is found.
In complex scenarios, the sampled waypoints will be densely distributed around the obstacle. To lower the computational cost, the density is reduced by a parameter $d_{\mathrm{insret}} > 0$. This means the sampling process in Fig. \ref{fig:ustc-inf} at a distance less than $d_{\mathrm{insret}}$ around other waypoints will not be performed. A new path is formed by combining the additional waypoints with the existing path. The above process is executed recursively until the generated path stops changing, indicating a global collision-free path is found.

\subsubsection{Trajectory Planner}
\label{sec:trajectory}

A time-dependent polynomial trajectory is generated by the trajectory planner module to connect path points using the minimum snap method with time allocation \cite{Richter2016}. In this method, the maximum flight speed allowed by the trajectory is set with $v_{\max}$. The optimization formulation is shown in Equation~\ref{eq:optimal}, where $c \in \mathbb{R}^{(k-1)(n+1)}$ contains the polynomial coefficients of the curve of order $n$, and $T_i$, $\gamma$ are the time duration of each segment and the weight of the total flight time cost. $A_{\text{eq}}$, $b_{\text{eq}}$ and $A_{\text{ineq}}$, $b_{\text{ineq}}$ represent the continuity constraints and the velocity constraints $v \leq v_{\max}$ of the trajectory respectively. The hyperparameters $v_{\max}$ and $\gamma$ can be adjusted to improve the overall planning result.

\begin{equation}
\begin{aligned}\label{eq:optimal}
	\min_{c,T_i}&		\ c^{\top}Qc+\gamma \sum_i{T_i}\\
	\,\,\text{s.t.} \ &		A_{\text{eq}}c=b_{\text{eq}}\\
	&		A_{\text{ineq}}c\leq b_{\text{ineq}},\\
\end{aligned}
\end{equation}

\subsubsection{Learning Based Compensation}
\label{sec:learning}

\begin{figure}
    \centering
    \includegraphics[scale=0.40]{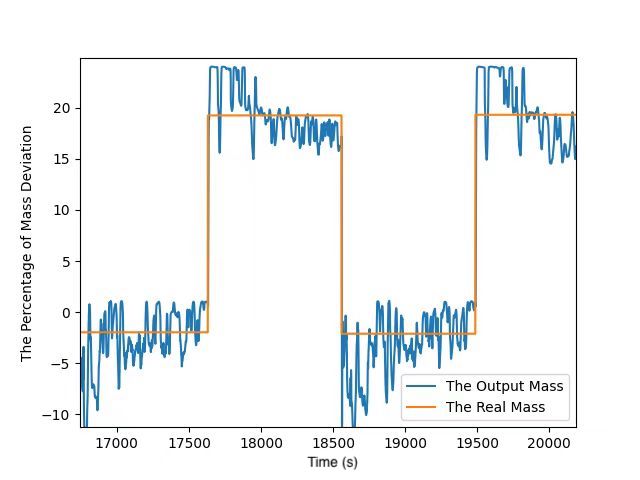}
    \caption{Mass error percentage estimation process of the LSTM. The percentage is calculated by $\frac{m-m_{\mathrm{nominal}}}{m_{\mathrm{nominal}}} \times 100\%$.}
    \label{fig:estimation}
\end{figure}

In the learning-based compensation module, a Long-Short Term Memory (LSTM) network is used to process a time series created from the quadrotor states and estimate the deviation of the quadrotor’s inertial parameters. 
We use the mean square error (MSE) as the loss function. The network’s input is set as the time sequence of z-axis coordinates, and the output is set as the percentage of mass deviation. In our analysis, we found that the rotational inertia bias has a relatively small impact on the actual flight performance and is difficult to compensate for by the reference signal due to the low command frequency. As a result, we only consider mass estimation and correction. The network runs in real time and outputs its estimation at every control step. The network contains $1$ input layer with $6$ nodes, $2$ hidden layers with $32$ nodes each, $2$ stacked layers, and $1$ output layer. Logs of 16 uncompensated flights are combined into one dataset for training, which has a total length of about 15000 control steps. The network parameters are updated once per step. Since the LSTM needs to receive a sufficient amount of data before it can produce a stable output, its outputs are stored in a first-in-first-out queue, so the estimation is available as long as the variance of the queue is less than a predefined threshold $\tau_{\mathrm{var}}$. Finally, the estimated mass bias $\Delta m$ is added to the quadrotor's kinematic equation for feed-forward compensation:
\begin{equation}
\begin{aligned}
a^*_{\mathrm{ref}} =\frac{1}{m} R \cdot f + \left( 1 + \frac{\Delta m}{m}\right) g  = a_{\mathrm{ref}} + \frac{\Delta m}{m}g,
\end{aligned}
\end{equation}
where $a_{\mathrm{ref}}$ and $a^*_{\mathrm{ref}}$ are the reference accelerations before and after compensation respectively, $m$ is the uncompensated mass. $f$ represents the thrust of the drone, $R$ represents the rotation matrix, and the acceleration of gravity is denoted as~$g$.
The estimation process is terminated after the network's output converges. 
Figure~\ref{fig:estimation} shows a validation process of the LSTM. 4 different flight logs with 2 kinds of mass deviations are used to test the network. The red line shows the real mass error percentage, and the blue line illustrates the estimated percentage.

\subsubsection{Dynamic Adjustment}
\label{sec:adjust}

In the dynamic adjustment module, an event-triggered position-based weighted summation method is used to dynamically modify the quadrotor's pre-planned positional reference $f_{\mathrm{ref}}(t)$. Here, a positional trajectory is denoted as a time-dependent 3D vector $f(t)=[x(t), y(t), z(t)]^T$. 
At the beginning of a flight, the nominal position $g_i$ of each gate $i$ is provided while the accurate position $g^*_i$ is unknown. When the quadrotor is sufficiently close to gate $i$, then $g^*_i$ is acquired. 
At this time, the dynamic adjustment mechanism is triggered, and the distance from the quadrotor’s current position to $g_i$ is calculated as $d_{i,0}$. 
From here on, until the quadrotor reaches the gate, the distance from its real-time position to $g_i$ is denoted as $d_{i}(t)$. Consequently, the real-time reference trajectory $f^*_{\mathrm{ref}}(t)$ can be generated using:
\begin{equation}
f^*_{\mathrm{ref}}(t) = f_{\mathrm{ref}}(t)+\left(1-\frac{d_{i}(t)}{ d_{i,0}}\right)(g^*_i - g_i) \quad d_{i}(t) \leq d_{i,0}.
\end{equation}

\subsection{Implementation}
\label{sec:ustc-implementation}

This project is based on the \texttt{\small safe-control-gym} framework provided by the competition. The open-source code\footnote{\url{https://github.com/ustc-arg/Safe-Robot-Learning-Competition}} is written in Python~3.8 and tested both on Ubuntu~18.04 and Windows~10 operating systems.

We developed multiple Python classes that have been built to implement specific functions matching the modules in the previous subsection.

As the competition includes four difficulty scenario settings---from Level $0$ to $3$---and all parameters in the solution must be tuned according to the different difficulty levels, to better manage and regulate these parameters, we created a separate Python class called parameter-assigner.
With this, we can automatically determine the current difficulty level and load the corresponding parameters into the other Python classes.

Next, we introduce the construction of the LSTM network. First, we performed several simulation tests without compensating the reference signals, in a collision-free environment. We found that the mass bias influences the z-axis height-tracking of the quadrotor most, leading to a steady-state error. After training, the LSTM can produce a reasonably accurate estimation within $2$ to $3$ seconds during most simulation tests.

\begin{figure}
    \centering
    \includegraphics[draft=False,width=.99\columnwidth, height=5cm]{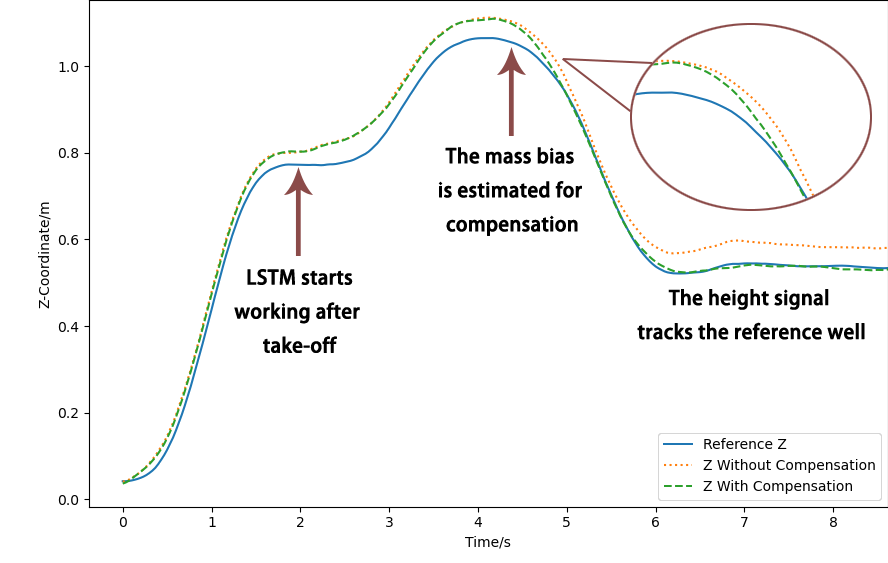}
    \caption{Height compensation process. The LSTM starts to estimate the mass deviation from the reference signal and the actual state after 2s. The estimation converges after about 4.5s and the output is used for feed-forward compensation.}
    \label{fig:ustc:compensation}
\end{figure}

The parameters introduced in \ref{sec:ustc-methods} are detailed at different difficulty levels in Table~\ref{table:ustc}. $d_{\mathrm{gate}}$ is tuned to $0.3$ in all difficulty levels, and other parameters are manually adjusted according to the simulation result. The parameter settings for the higher difficulty levels generate more conservative results.

\begin{table}
    \centering
    \caption{Parameter Settings}
    \begin{tabular}{ccccc}
    \toprule
      & level 0 & level 1 & level 2 & level 3 \\
        \cmidrule(lr){1-5}
        $d_{\mathrm{gate}}$ & 0.3 & 0.3 & 0.3 & 0.3 \\
    $\alpha$ & $\pi/3$ & $\pi/4$ & $\pi/4$ & $11\pi/36$ \\
    $d_{\mathrm{insret}}$ & 0.3 & 0.2 & 0.2 & 0.2 \\
    $v_{\max}$ & 4 & 3 & 3 & 2 \\
    $\gamma$ & 25000 & 500 & 500 & 25000 \\
    $\tau_{\mathrm{var}}$ & 0 & 0 & 0.05 & 0.05 \\
    \bottomrule
\end{tabular}     \label{table:ustc}
\end{table}

\subsection{Results}
\label{sec:ustc-results}

The associated video is available \href{https://youtu.be/kVlCnwWd_Z8}{here}.
Having finely tuned the parameters, team USTC's solution is able to achieve the goals required by the competition task. According to the simulation results, the total completion time of Level 0 and Level 1 scenarios is about 9.7 seconds and increases to 11.6-12.5 seconds for Level 2 and Level 3. While still successful, the completion time increased to 15.9 seconds for real-world flight tests. Nonetheless, the solution performed well enough to earn the competition's 3rd place.

It is also important to point out an open question about a safety-performance trade-off that was faced by the developer of the solution. In the planning phase, using more aggressive parameters reduces the success rate of the task, while using more conservative parameters increases the completion time and reduces the competitiveness of the results. 
 \section{Team Ekuflie's Solution}
\label{sec:ekuflie}

The participation of Team Ekuflie in this competition was a technology exploration
project by the Research and Development (R\&D) division at Ekumen Labs, a robotics software
development company.
This R\&D division's purpose is to expand the company's know-how and better understand both the limitations and promises of state-of-the-art robotics technology.

Team Ekuflie started the competition with limited background in optimal control theory, but with
ample experience in higher-level applications of navigation, planning, perception, and other related
robotics technologies.

\subsection{Methods}
\label{sec:eku-methods}

A number of different approaches, both control and learning-based, were initially investigated.
This stage culminated with the selection of a model predictive contouring control (MPCC) methodology based on~\cite{romero2021} combined with the time-optimal online replanning described in~\cite{romero2022}.
This combination suited several of the goals of the competition:
\begin{itemize}
    \item Minimizing transit time is a first-order optimization goal of the algorithm.
    \item Being robust against model imperfections and perturbations through online path replanning.
    \item Allowing to update the gates' pose information during execution.
\end{itemize}

The following subsections outline MPCC and time-optimal online replanning.
Ekuflie's custom modifications to the base algorithm---to account for competition requirements and performance constraints---are enumerated later in Subsection~\ref{sec:eku-implementation}.

\subsubsection{MPCC}
\label{sec:mpcc}

MPCC is an optimal control strategy based on model predictive control (MPC). 
The goal of MPC is to track a discrete-time reference trajectory as closely as possible. 
In contrast, MPCC accepts a differentiable curve $p^d$ through space as a reference and tries to
minimize the Euclidean distance to the curve, while also maximizing the progress along it.

At every time step, MPCC solves a constrained optimization problem over a finite number of future time steps, the horizon~$N$. 
MPCC relies on a model to predict the system's state at future time steps.  
Consider the nonlinear discrete-time dynamics model describing a quadrotor's motion
$\vec{x}^d_{k + 1}= 
f(\vec{x}^d_k, 
\vec{u}^d_k
)$, where $\vec{x}^d_k$ is the aggregation of the quadrotor states for
position $\vec{p}$, rotation $\vec{q}$, linear velocity $\vec{v}$, and body rates $\vec{w}$, while
$\vec{u}^d_k$
is the control input vector containing the individual rotor thrusts.
The MPCC algorithm presented in~\cite{romero2022} augments this dynamics model as follows:
\begin{align*}
    \vec{x}^d_{k+1} &=  f(\vec{x}^d_k,
    \vec{u}^d_k
    ) \notag \,, \\
    \vec{u}^d
    _{k+1} &= 
    \vec{u}^d_k + \Delta 
    \vec{u}^d_k
    \Delta t \notag \,, \\
    \theta_{k+1} &= \theta_k + v_{\theta, k} \Delta t \notag \,, \\
    v_{\theta, k+1} &= v_{\theta, k} + \Delta v_{\theta, k} \Delta t \,,
\end{align*}
where $\theta$ is the progress along the reference path (i.e. arc-length from the start) and $v_{\theta}$ is the progress' rate of change (i.e. velocity
along the reference curve). The rates $\Delta u^d$ and $\Delta v_\theta$ are also introduced.
The purpose of this augmentation is twofold. 
First, the explicit expression of the progress and its rate enables their inclusion in the optimization problem, e.g. maximizing the
progress along the reference curve. Second, constraints can be formulated on the additional rates, e.g. to improve the
optimization problems' stability, see~\cite{romero2021}.

The augmented dynamics model can be written as $x_{k + 1} = g(x_k, u_k)$ with augmented state and input vectors:
\begin{align*}
    \vec{x} = 
    \left[
    \vec{p}^\transpose\,, \vec{q}^\transpose\,, \vec{v}^\transpose\,, \vec{w}^\transpose \,, 
    \vec{u}^{d\transpose}
    \,, \theta \,, v_{\theta} 
    \right]
    ^\transpose, 
    \vec{u} = 
    \left[
    \Delta v_{\theta} \,, \Delta
    \vec{u}^{d\transpose}
    \right]
    ^\transpose \,.
\end{align*}
The augmented state vector is a combination of the original quadrotor model states and a number of virtual states added through the additional state equations
present in the augmented system: 
$\vec{u}^d$, $\theta$ and $v_{\theta}$.

The optimization problem's constraints are given by the augmented dynamics model $x_{k + 1} = g(\vec{x}_k, \vec{u}_k)$ as
an equality constraint, along with inequality constraints~(box constraints) on the magnitude of several state variables
(see Eq. 17 in~\cite{romero2021}).

\begin{figure}
  \centering
  \includegraphics[width=0.95\columnwidth,keepaspectratio]
  {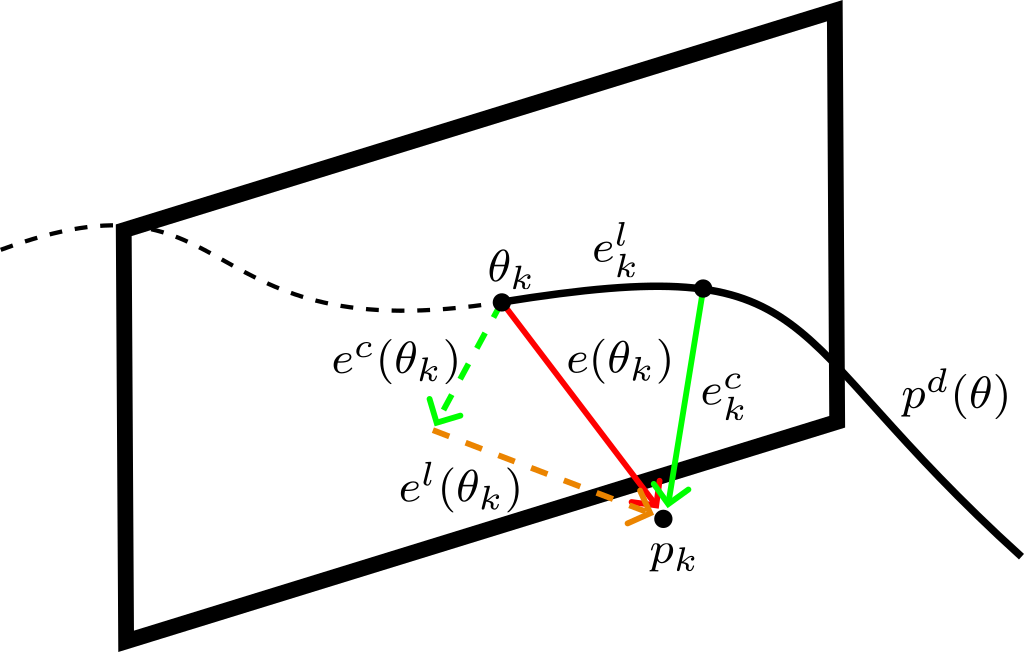}
  \caption{
  Visualization of the reference curve $p^d$, the quadrotor's position $p_k$ at time step $k$, and the tracking errors.
  The approximated lag and contour tracking errors, $e^l_k(\theta_k)$ and $e^c_k(\theta_k)$, respectively,
  are defined as the components of $e(\theta_k)$ that are
  normal and parallel, respectively, to the normal plane at $\theta_k$. 
  These approximations estimate the true lag and contour errors $e^l$ and $e^c$, respectively, see~\cite{romero2021} for details.  
}
  \label{fig:lag_contour_errors}
\end{figure}

The optimization problem's cost function includes terms for the contouring objectives, e.g. the minimization of contouring errors to stay close to the reference.
The lag and contour tracking errors, $e^l_k$ and $e^c_k$, respectively, are determined with respect to the quadrotor's current position $p_k$ and its approximated position on the reference curve $p^d(\theta_k)$, see~\autoref{fig:lag_contour_errors} for definitions.
To achieve real-time online optimization, these tracking errors need to be approximated as $e^l(\theta_k)$ and $e^c(\theta_k)$, respectively.
Additional cost function terms minimize the quadrotor's body rates for improved quadrotor stability and the augmented control inputs for minimal control effort. 
A final term with negative cost weight $-\mu$ encourages the maximization of the progress rate $v_{\theta}$.
The result of the optimization is a sequence of control inputs $\vec{u}_{k + i}^* \,, \forall i \in \{0, \dots, N - 1 \}$
that results in a minimum-cost trajectory according to the dynamics model used in the derivation.
As in MPC, only the first control input $u_{k}^*$ from the sequence is applied to the quadrotor at each control loop iteration, and the MPCC's optimization problem is solved again at the next time step in a receding horizon fashion.

\subsubsection{Optimal reference replanning}
\label{sec:replan}

A differentiable reference curve for MPCC can be obtained in various ways.
The chosen approach was proposed in~\cite{romero2021} and is
a computationally inexpensive solution that generates a smooth 3D curve through
a series of waypoints using a simple particle-mass model. Furthermore, the authors in~\cite{romero2021} show that using this method for MPCC
yields performances comparable to that of a much more computationally
expensive full-model trajectory optimization.

Given a known initial point-mass position $p_0$, velocity $v_0$, and assuming a point-mass dynamic such that $\ddot{p} = a$ with
acceleration values constrained by $\underline{a} \leq a \leq \overline{a}$,
it can be shown that the time-optimal control input to reach a target position $p_1$ and
velocity $v_1$ is to apply a bang-bang policy of the form:
\begin{align*} \label{eq:pmm_bang_bang}
  a(t) &= \begin{cases}
    \underline{a}, & 0 \leq t \leq t_1 \\
    \overline{a}, & t_1 \leq t \leq T \,,
  \end{cases}
\end{align*}
or vice versa, starting with $\overline{a}$, for some minimal time $T = t_1 + t_2$. This piecewise constant acceleration trajectory can be computed in closed form, for each consecutive pair of waypoints.
Provided that the waypoint positions are available, the algorithm proposed in~\cite{romero2021} samples waypoint velocities and recursively builds a so-called \textit{velocity graph}, weighted by
trajectory traversal times. An approximate time-optimal trajectory can then be found by standard shortest-path algorithms.

\subsection{Implementation}
\label{sec:eku-implementation}

The MPCC implementation in Team Ekuflie's competition submission\footnote{\url{https://github.com/ekumenlabs/safe-control-gym}} leverages CasADi's~\cite{Andersson2019} expression graphs and heavily relies on Just-In-Time compilation for performance. 
The quadrotor dynamics are modeled after~\cite{luis2016} and are discretized using an explicit second-order Runge-Kutta method. B-splines embody the differentiable, arc-length parameterized curve to be traced by the MPCC.
IPOPT is used to solve the MPCC's nonlinear optimization problem and is warm-started on every step using the state prediction from the
previous step.

The time-optimal path planning is implemented using minimal SciPy~\cite{scipy2020} and native Python, resorting to closed-form analytical
expressions (polynomial root finding, arc-length integrals, etc. ) whenever possible. NetworkX~\cite{networkx2008} provides the shortest
path algorithm.
The resulting curve finally undergoes arc-length reparameterization using an efficient technique described in~\cite{arclengthparameterizationwang2002}. This
creates a function $p(\theta)$ that represents the trajectory of
the curve through space and whose argument $\theta$ is the arc length from the initial position~$p_0$.

In addition to the baseline implementations for both controller and planner, Ekuflie's solution accounts for the competition-specific requirements and constraints. For convenience, we define the Euclidean distance from the reference position $p(\theta)$ to the $i$-th gate's position $p_{g, i}$ as:
\begin{equation*}
   \Delta p_i (\theta) =  \Vert p(\theta) - p_{g, i} \Vert \,.
\end{equation*}

  For safe gate traversal, the balance between maximizing progress and minimizing contour error is shifted
  towards contour error minimization via dynamic weight allocation (as proposed in \cite{romero2021}) when in
  proximity to the $i\mbox{-th}$ gate (out of $G$ gates in total):
  \begin{align*}
    q_c(p(\theta)) = \underline{q_c} + (\overline{q_c} - \underline{q_c}) \sum_{i = 1}^G \exp{\left(-\frac{\Delta p_i^2 (\theta)}{2 \sigma_e^2}\right)} \,,
  \end{align*}
  where $\underline{q_c}$ is the baseline contour error cost,
  and $\overline{q_c}$ is the peak cost in proximity of the $i$-th gate. The parameter $\sigma_e$ allows to adjust the
  width of the Gaussian function, which in turn determines the size of the area around the gate where the cost is increased.

  For obstacle avoidance, conservative path planning prunes all trajectories that may result in a collision considering obstacle
  pose distributions. Dynamic contour weight allocation includes the closest locations to each obstacle along the resulting time-optimal trajectory.

  Since time-optimal path planning is too computationally expensive to afford real-time replanning, it is performed only once at
  the beginning of each episode. This yields a sufficient approximation to the optimal path given that the location of all the obstacles
  is known (within some bounded uncertainty) at the start of the episode.

  To cope with inexact gate positions up until the quadrotor is in gate proximity, approaching gates faster than a given soft limit on $v_{\theta}$ is discouraged, ensuring
  corrective maneuvers are possible. This is encoded as a modified progress incentive factor $\mu_v$ in the cost function:
  \begin{align*}
    \mu_v =&  \mu \left[ 1 - U_{i}
    \cdot \exp{\left(1 + K_b \frac{v_{\theta} - \overline{v_b}}{\overline {v_b}}\right)} \right. \\
    &  \cdot \left.  \exp{\left(-\frac{\Delta p_i^2 (\theta)}{2 \sigma_b^2} \right) } \right] _{i=g}.
  \end{align*}
  In this expression $U_{i}$ is a binary variable encoding the uncertainty in the $i$-th gate location
  ($U_{i}=0$ if the pose is exactly known, $U_{i}=1$ otherwise).
  $K_b$ and $\overline{v_b}$ characterize a soft limit on $v_{\theta}$, while $\sigma_b$ determines the distance to
  the gate where the soft speed limit gradually begins to become effective. Finally, $g$ represents the index of the next gate to cross,

  Once exact gate positions are provided, and since full trajectory replanning cannot be conducted in real-time,
  simpler trajectory corrections $\delta_{p_{g}}(\theta)$ are applied on top of the nominal path planned at the start of the episode:
  \begin{align*}
    \delta_{p}(\theta) =& \left[ \delta_{p_{i}} \exp\left({-\frac{\Delta p_i^2 (\theta)}{2 \sigma_c^2}}\right) \right] _{i=g} + \\
    & \left[ \delta_{p_{i}} \exp\left({-\frac{\Delta p_i^2 (\theta)}{2 \sigma_c^2}}\right) \right] _{i=l} \,,
    \end{align*}
  where $\delta_{p_{i}}$ is the difference between the exact and approximate location of the $i$-th gate (or $0$ if the exact
  location is not yet known), and $\sigma_c$ controls the size of the exponential factor that gradually applies
  the correction when in proximity to the gate. Lastly, $l$ and $g$ represent the indices of the last gate that
  was crossed and the next one respectively.

  Since the control interface was not direct motor inputs but through an on-board Mellinger controller, Ekuflie's solution uses predicted kinematic states (ie. position, orientation, velocities)
  that are used as setpoints rather than predicted thrust forces. Having a Mellinger controller in the loop restricts the MPCC controller's ability to shape the system
  response and thus no attempt was made to learn and correct for uncertainty or bias in model dynamics. The Mellinger controller is also relied on for take-off, hard brake, and
  landing.

  In case of midflight MPCC convergence issues, a low-speed trajectory following controller was introduced as a fallback to keep the quadrotor airborne until MPCC recovery.

  Since the provided input observations lack information about the current velocity and body rates of the quadrotor, first-order linear velocity estimates were used to supplement the input measurements before being passed to the MPCC controller.

\subsection{Results}
\label{sec:eku-results}

\begin{figure}
  \centering
  \includegraphics[width=.99\columnwidth,keepaspectratio]{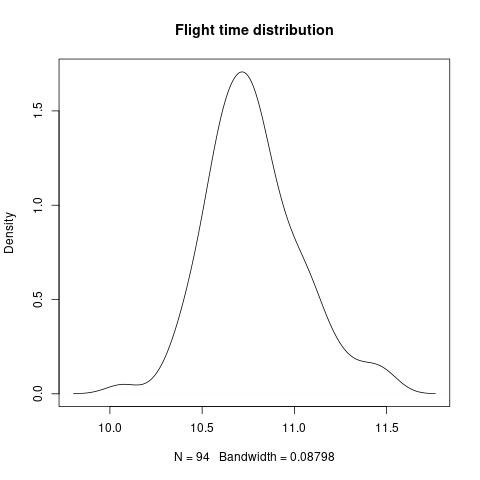}
  \caption{Flight time distribution for the Ekuflie team's solution estimated from 100 independent realizations of the `level 3` simulation configuration.}
  \label{fig:ekuflie_flight_time}
\end{figure}

The performance of the complete solution was analyzed by running a hundred different realizations of
the \textit{level 3} simulation configuration. This is the most demanding setting of the challenge,
where both the inertial properties of the quadrotor and the poses of the obstacles are randomized, and where
the random seed is not reset between episodes to ensure independence between different realizations of the same simulation.

During this analysis, collisions were observed at a $2\%$ rate. Constraint violations were observed at a $4\%$ rate, mostly due
to initial state randomization taking the quadrotor too close to workspace boundaries. MPCC convergence issues
were observed at a $6\%$ rate. A flight time distribution estimate is shown in Figure~\ref{fig:ekuflie_flight_time},
ranging between $10 \mbox{s}$ and $11.5 \mbox{s}$ and where a non-negligible fraction of the total time is explained by take off,
hard brake, and landing.

MPCC is thus shown to be a valid control strategy for nano-quadrotors, with no prior training phase. However, these time-optimal
trajectory planning and MPCC implementations require substantial computing resources. Arguably, neither CPython nor
IPOPT lend themselves well to real-time applications. Trajectory corrections are a brittle substitute for online
replanning. Cost shaping for speed limiting makes controller tuning harder than it already is.

Team Ekuflie's solution was awarded second place during the actual competition. A recording of the performance
both in simulation and during the sim2real round can be seen in the \href{https://www.youtube.com/watch?v=C6PZYJ5R1MI}{competition summary video}.
While the solution performed as expected in levels 0 to 3 (simulation), the time performance in the sim2real scenario was below the
equivalent simulated scenario; while the causes for this still need further investigation, this was most likely due to the control loop rate failing
to meet its nominal rate due to the latencies in the system (MPCC controller calculation, radio messaging, etc.).

Moving forward, these implementations need to be migrated to a suitable computing platform that can meet the
required control rates. Having full state observations and dropping the Mellinger controller would also allow the MPCC
controller to perform as intended with no restrictions. It also remains to be seen if point-mass time-optimal
trajectories are a good approximation for low thrust-to-weight ratio quadrotors.
 \section{Team H$^2$'s Solution}
\label{sec:h2}

Team H$^2$ was formed by a group of 3 robotics engineers from Singapore who were interested in learning more about planning and control algorithms on aerial robotic systems. The team members had experience in applications of navigation, planning, and control algorithms on land and underwater systems, and set out to learn how these could be applied to the drone with this competition.

As the competition environment has well-defined uncertainty bounds for the poses of the drone, obstacles, and the gates, specified by known probability distributions, team H$^2$ adopted an approach based on traditional path and trajectory planning for the solution.

\begin{figure}\centering
    \begin{tikzpicture}
    \begin{groupplot}[
    group style = {group size = 1 by 1, horizontal sep=1.5cm, vertical sep=1.5cm,},
    width = 1.0\columnwidth, 
    height = 5.0cm,
    xmin=0, 
    xmax=100, 
    ymin=0, 
    ymax=100,
grid=none, axis line style={draw=none}, 
    tick style={draw=none}, 
    clip=false,
    xticklabel=\empty, 
    yticklabel=\empty,
    axis on top=true,
    clip marker paths=true
    ]

    \nextgroupplot[]
        \node[opacity=1.0] at (axis cs:50,50) () {\includegraphics[width=.99\columnwidth]{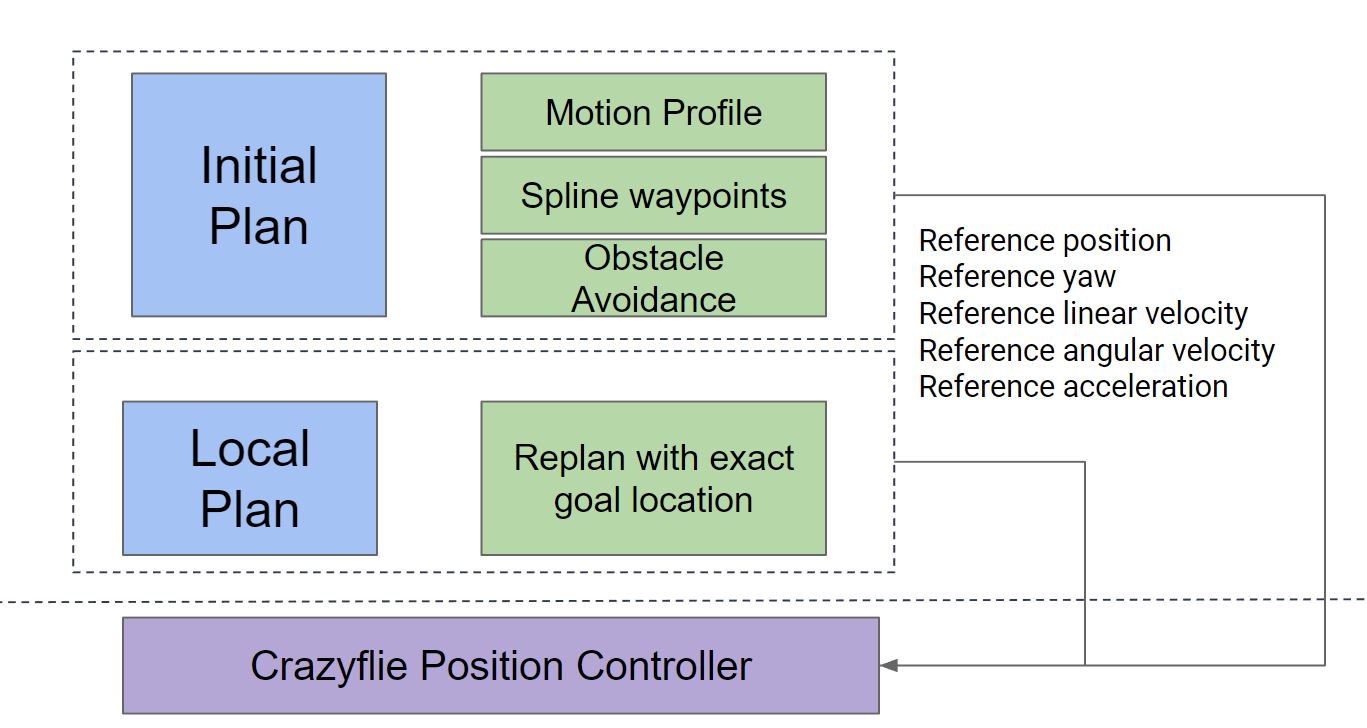}};
    \end{groupplot}

\end{tikzpicture}
\vspace{-1em}
     \caption{Architecture of Team H$^2$'s proposed solution.}
    \label{fig:h2_sol}
\end{figure}

The proposed solution (Figure~{\ref{fig:h2_sol}}) comprises 2 layers: an initial plan and a local re-planner. The initial plan is based on the a priori information about gate positions, as well as obstacle locations. As the initial locations are only approximate and the exact positions become known to the drone later, a local re-planner is needed to ensure that the trajectory of the drone passed through the gate. There were 3 main tasks within the initial plan: 
\begin{enumerate}
    \item generating spline coefficients for each path segment between waypoints;
    \item selecting a motion profile;
    \item and obstacle avoidance.
\end{enumerate}

The full state commands of the initial plan are sent to the drone until the exact location of the nearest upcoming gate becomes known. At this point, the local layer performs a re-planning, by generating a path from the current state of the drone to the gate, and from the gate to the next waypoint.

\subsection{Methods}
\label{sec:h2-methods}

For each of the sub-problems, there are several considerations to make, with the main concerns being the safety of the robot, computation time, and the time taken for the drone to complete the course.

\subsubsection{Waypoint Splines}
\label{sec:h2-splines}

The first step in the solution is to create a spline joining all the waypoints. Cubic and quintic polynomials were considered for the x, y, and yaw axes, but quintic polynomials were eventually chosen as they are continuous in position, velocity, and acceleration. There were 2 possible heights for the competition's gates, which implies that the waypoints on the z-axes would take 2 possible discrete values. Hence, a Piecewise Cubic Hermite Interpolating Polynomial (PCHIP) was initially chosen for the z-axis. However, the discontinuity in accelerations resulted in oscillations in the z-axis, and quintic polynomials were eventually used for it as well. As we would like the drone to always face the direction of travel,
the yaw of the drone is the tangent of the trajectory. The x and y velocities at each waypoint are the constraints for the quintic polynomial on the yaw axes. The linear acceleration along the x and y axes corresponds to the yaw angular velocity. The yaw rate at gates was chosen to be 0 rad/s, which resulted in better tracking performance. This fully constrains the quintic polynomials used to describe the position of the drone in each dimension (x,y,z) in time.

\begin{equation}
    x(t) = a_0 + a_1t + a_2t^2 +  a_3t^3 + a_4t^4 + a_5t^5
\end{equation}

\begin{equation}
    \begin{split}
        a_5 &= x_0 \\
        a_4 &= \dot{x}_0 \\
        a_3 &= \frac{\ddot{x}_0}{2} \\
        a_2 &= \frac{20(x_f-x_0)-(8\dot{x}_f+12\dot{x}_0)t_f+(\ddot{x}_f-3\ddot{x}_0)t_f^2}{2t_f^3} \\
        a_1 &= \frac{-30(x_f-x_0)+(14\dot{x}_f+16\dot{x}_0)t_f-(2\ddot{x}_f-3\ddot{x}_0)t_f^2}{2t_f^4} \\
        a_0 &= \frac{12(x_f-x_0)-6(\dot{x}_f+\dot{x}_0)t_f+(\ddot{x}_f-\ddot{x}_0)t_f^2}{2t_f^5}
    \end{split}
\end{equation}

The term $a_n$ refers to the coefficient of the polynomial for the $t^n$ term. $x_0$ and $x_f$ are the starting and final position, while $t_f$ is the final time. 

\subsubsection{Motion Profile}
\label{sec:h2-motion}

Having determined the shape of the path passing through all the waypoints, the next step is to design a motion profile for the drone to traverse the path. We considered 2 types of motion profiles---S-curve and polynomial. The S-curve motion profile is a blend of cubic polynomials with linear sections, while the polynomial motion profile would simply be a quintic polynomial.

\label{sec:h2-s-curve}
\begin{figure}[hbt!]
    \centering
    \begin{tikzpicture}
    \begin{groupplot}[
    group style = {group size = 1 by 1, horizontal sep=1.5cm, vertical sep=1.5cm,},
    width = 1.0\columnwidth, 
    height = 6.5cm,
    xmin=0, 
    xmax=100, 
    ymin=0, 
    ymax=100,
grid=none, axis line style={draw=none}, 
    tick style={draw=none}, 
    clip=false,
    xticklabel=\empty, 
    yticklabel=\empty,
    axis on top=true,
    clip marker paths=true
    ]

    \nextgroupplot[]
        \node[opacity=1.0] at (axis cs:50,50) () {\includegraphics[width=.99\columnwidth, 
]{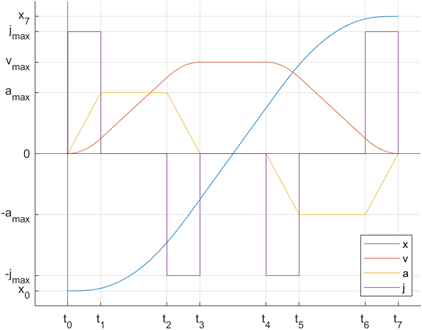}};
    \end{groupplot}

\end{tikzpicture}
\vspace{-1.5em}     \caption{S-curve.}
    \label{fig:h2_s_curve}
\end{figure}

\begin{figure}[hbt!]
    \centering
    \begin{tikzpicture}
    \begin{groupplot}[
    group style = {group size = 1 by 1, horizontal sep=1.5cm, vertical sep=1.5cm,},
    width = 1.0\columnwidth, 
    height = 4.0cm,
    xmin=0, 
    xmax=100, 
    ymin=0, 
    ymax=100,
grid=none, axis line style={draw=none}, 
    tick style={draw=none}, 
    clip=false,
    xticklabel=\empty, 
    yticklabel=\empty,
    axis on top=true,
]

    \nextgroupplot[]
        \node[opacity=1.0] at (axis cs:50,50) () {\includegraphics[width=.99\columnwidth, 
]{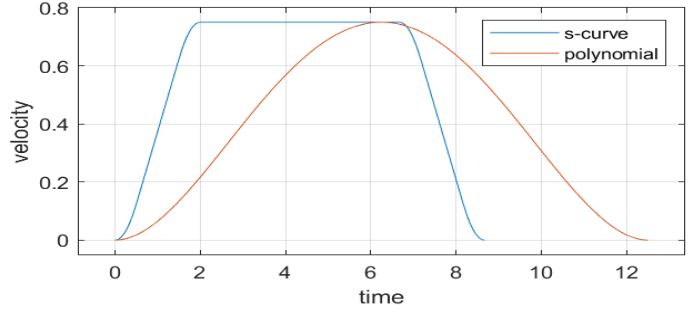}};
    \end{groupplot}

\end{tikzpicture}
\vspace{-2em}     \caption{Velocity Motion Profile.}
    \label{fig:h2_motion_profile}
\end{figure}

We see from Figure~\ref{fig:h2_motion_profile} that, for the same desired displacement (area under velocity curve), the drone travels for a longer duration at the maximum velocity with the s-curve motion profile, leading to faster completion times. Hence, the S-curve motion profile was selected for H$^2$'s solution.

\label{sec:h2-motion-profile}

\subsubsection{Obstacle Avoidance}
\label{sec:h2-obstacle}

The minimum allowable distance \emph{r} away from an obstacle is given by:
\begin{equation}
    r = r_{obs} + r_{tol}
\end{equation}
where $r_{obs}$ is the radius of the obstacle, and $r_{tol}$ is an inflation radius around the obstacle, which can be tuned proportionally to the desired clearances of the planned trajectory from the obstacles. With knowledge of the polynomial coefficients at each path segment and the position of the obstacles, a closed-form polynomial function was used to describe the distance of the path from each obstacle for each path segment. The equation for the intersection between the trajectory and the inflation radius is as follows:
\begin{multline} \label{eq:1}
    0 = {{(a_0 + a_1 \,t + a_2 \,t^2 + a_3 \,t^3 + a_4 \,t^4 + a_5 \,t^5 - x_0)}}^2 +\\
    {{(b_0 + b_1 \,t + b_2 \,t^2 + b_3 \,t^3 + b_4 \,t^4 + b_5 \,t^5 - y_0)}}^2 - r^2,\\
    \forall \ t\in(t_0,t_f)
\end{multline}
where $a_k$ and $b_k$ are the coefficients of the $x$ and $y$ axis polynomial respectively, and the obstacle is located at $(x_0,y_0)$ with inflation radius \emph{r}. The intersection between the planned path and the inflation circle was found by solving for the roots of \emph{t} in the 10\textsuperscript{th} order polynomial Equation~\eqref{eq:1}. If real roots exist within the parameter interval, the trajectory intersects the inflation radius around the obstacle. 

A new waypoint was created by finding the point on the trajectory with minimum distance to the circle center, and projecting it outwards to the distance \emph{r} away from the obstacle. The trajectory segment is then re-planned with two segments, one from the previous waypoint to the new point, and another from this new point to the next waypoint. Although this method does not guarantee that the new trajectory lies outside of the inflation regions, collisions could be avoided as long as $r_{tol}$ is sufficient and obstacles are sparse.

\subsubsection{Local Re-planning}
\label{sec:h2-replan}

The local re-planning layer was needed as the positions of the gate used for the initial planning were approximate, and the exact location of the gates would only be updated as the drone approached the gate. When the exact location of the gate becomes known, quintic polynomial segments are added from the drone's current position to the true location of the gate, and from this new gate location to the next waypoint. The current position, velocity, and acceleration of the drone are the constraints for computing polynomial coefficients of the added path segment.

\subsection{Implementation}
\label{sec:h2-implementation}

With the approach for the solution mapped out in the previous section, there were some additional practical adaptations\footnote{\url{https://github.com/huiyulhy/safe-control-gym/tree/lspb_s_curve}}  needed in the algorithms to enable the drone to be able to track the trajectory created from the waypoint splines with the selected motion profile.

\subsubsection{Gate orientation}
\label{sec:h2-gate}

When creating the polynomial function joining all waypoints, the gates could have 2 possible orientations---in the direction of the start to end of the path and in the opposite direction. As the orientation of the gate is used as a velocity constraint for creating the drone's trajectory, it was important to ensure that the gate was oriented in the direction from start to end of the path, for the shortest path to be generated. A check based on the vector dot product of the current gate heading with the displacement vector of the previous and next gate was added to determine if the gate orientation needed to be flipped.

\subsubsection{Waypoint Splines}
\label{sec:h2-splines2}

After determining the final position of the gates to be used for creating the waypoint splines, quintic polynomials were used for the x, y, and z axes. The polynomial function describing the drone's trajectory on the x-axis is:
\begin{multline}
    x_{i}(t) = a_{0,i} + a_{1, i}t_{i} + a_{2, i}t_{i}^2 +  a_{3, i}t_{i}^3 + a_{4, i}t_{i}^4 + a_{5, i}t_{i}^5 \\
    \forall i \in \{0, 1, \cdots, N\} ,  t_{i, 0} \leq t \leq t_{i, f}
\end{multline}
where $N$ is the number of waypoints the drone needs to pass through and $t_{i}=t-t_{i, 0}$. The same form was used to describe the drone's position and orientation on the y and z axes. 

To use this formula, one needs to know all $t_{i, 0}$ and $t_{i, f}$. An initial estimate of the end time for each path segment was computed as a function of the Euclidean distance between the start and endpoints of the path segment, and the maximum velocity of the drone. This gave a lower bound of the distance of the path segment between the 2 waypoints,  which introduced regions in the trajectory where the motion profile could exceed kinematic limits. However, this is addressed with a time-scaling algorithm further discussed below.

\subsubsection{Motion Profile}
\label{sec:h2-motion2}

\begin{figure}[hbt!]
    \centering
    \begin{tikzpicture}
    \begin{groupplot}[
    group style = {group size = 1 by 1, horizontal sep=1.5cm, vertical sep=1.5cm,},
    width = 1.0\columnwidth, 
    height = 4.0cm,
    xmin=0, 
    xmax=100, 
    ymin=0, 
    ymax=100,
grid=none, axis line style={draw=none}, 
    tick style={draw=none}, 
    clip=false,
    xticklabel=\empty, 
    yticklabel=\empty,
    axis on top=true,
    clip marker paths=true
    ]

    \nextgroupplot[]
        \node[opacity=1.0] at (axis cs:50,50) () {\includegraphics[width=.99\columnwidth, 
        height=3cm
        ]{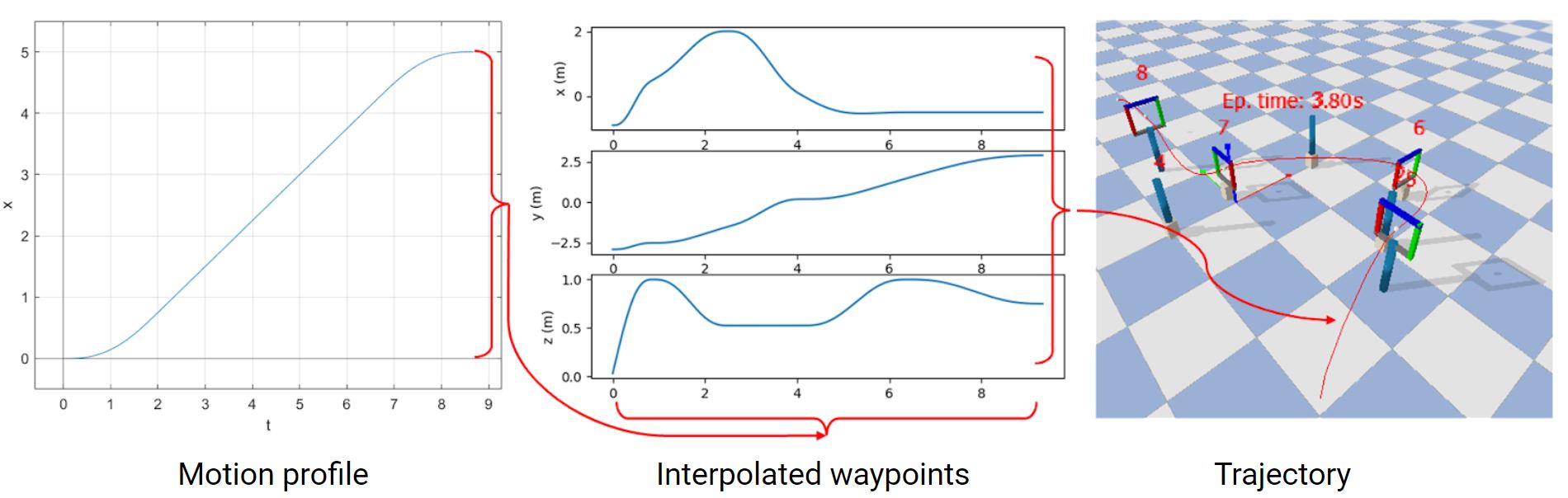}};
    \end{groupplot}

\end{tikzpicture}
\vspace{-1em}     \caption{Mapping from the motion profile onto the interpolated path.}
    \label{fig:h2_parameterization}
\end{figure}

The output of the motion profile is mapped onto the interpolated path, by scaling the distance traveled along the motion profile by the total path length for each path segment. As the $x, y, z$ positions are described by separate polynomials while the motion profile is a single curve---to couple velocity, acceleration, and jerk limits across all axes---the motion profile was parameterized so that each point along the motion profile mapped to a point on the $x, y, z$ trajectory. Figure~\ref{fig:h2_parameterization} shows the mapping from the motion profile onto the interpolated path. The following constraint describes the relationship between the points in the motion profile and the interpolated path:
\begin{equation}
    \delta s^2 = \delta x^2 + \delta y^2 + \delta z^2
\end{equation}
where $\delta s$ is an infinitesimal segment along the motion profile, and $\delta x, \delta y, \delta z$ are infinitesimal segments along the $x$, $y$ and $z$ axes respectively.

\subsubsection{Trajectory Tracking and Adaptation}
\label{sec:h2-adaptation}

Mapping from the motion profile to the interpolated path produces a desired position and yaw, linear and angular velocity, and linear acceleration for trajectory tracking. However, there were several sources of errors that could affect the performance of the drone's trajectory, mainly being: 
\begin{itemize}
    \item uncertainty in the kinematic limits;
    \item uncertainty in model parameters (e.g. weight, etc.);
    \item and disturbances.
\end{itemize}

\begin{figure}[hbt!]
    \centering
    \begin{tikzpicture}
    \begin{groupplot}[
    group style = {group size = 1 by 1, horizontal sep=1.5cm, vertical sep=1.5cm,},
    width = 1.0\columnwidth, 
    height = 4.0cm,
    xmin=0, 
    xmax=100, 
    ymin=0, 
    ymax=100,
grid=none, axis line style={draw=none}, 
    tick style={draw=none}, 
    clip=false,
    xticklabel=\empty, 
    yticklabel=\empty,
    axis on top=true,
    clip marker paths=true
    ]

    \nextgroupplot[]
        \node[opacity=1.0] at (axis cs:50,50) () {\includegraphics[width=.99\columnwidth, 
]{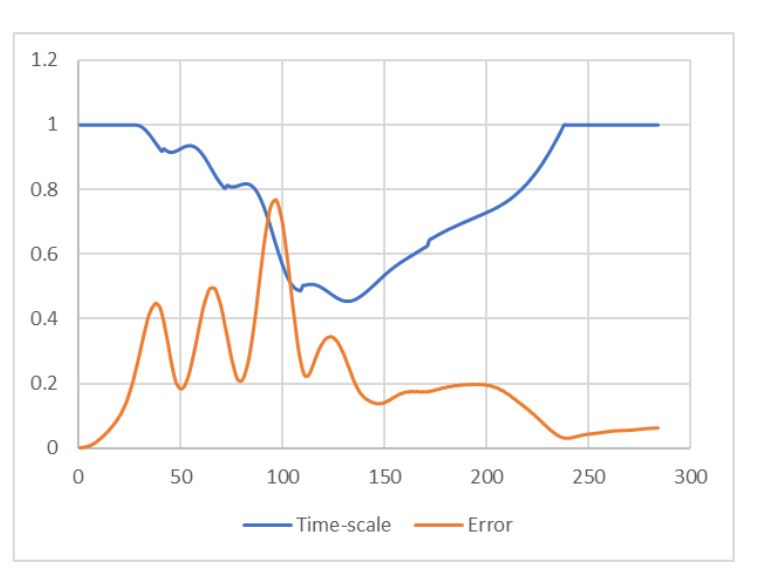}};
    \end{groupplot}

\end{tikzpicture}
\vspace{-1em}     \caption{Example of time scale varying with trajectory error}
    \label{fig:h2_time_scale}
\end{figure}

These can lead to the build-up of tracking errors in the trajectory. Thus, a time-scaling factor was introduced to improve path tracking. This was in the form of:
\begin{equation}
\tau = a \log( \|e\| + b), 0.0 \leq \tau \leq 1.0
\end{equation}
where $\tau$ refers to the time-scale factor used, and $a$ and $b$ were hyperparameters to be tuned. When tracking errors were low, $\tau$ tends towards $a\log b$, and at higher tracking errors, $\tau$ tends towards $a \log e$.  This then lengthens the trajectory, until the tracking error reduces and allows the time scale to return towards 1.0.  
Figure~\ref{fig:h2_time_scale} shows an example of how the timescale works as a function of the trajectory error. 

\subsubsection{Full State Commands}
\label{sec:h2-commands}

The outputs from the mapping of the motion profile to the interpolated path and the local re-planning layer are the reference position and yaw, linear and angular velocity, and linear acceleration commands to the position controller onboard the Bitcraze Crazieflie. Small angle assumptions were made when calculating the reference angular velocities in roll and pitch.

\begin{equation}
    \begin{split}
        \dot{v}_x &\simeq g\theta \\
        \dot{v}_y &\simeq g\phi
    \end{split}
\end{equation}

\subsection{Results}
\label{sec:h2-results}

The solution was tested in the competition simulation environment, with the levels provided by the competition code-base, as well as levels with custom positions of gates and obstacles to test the robustness of the algorithms. A total of 10 different scenarios were used, with varying obstacle locations and gate positions and orientations. The gate and obstacle positions and orientations were varied such that all of them would remain within the boundaries of the physical environment. Each scenario was run thrice to ensure repeatability in results. The team used several metrics to measure safety and efficiency (Table~\ref{tab:h2_metrics}). During the testing, the simulation terminated upon collision or constraint violation. 

\begin{table}[ht]
\centering
\caption{Performance and Safety Metrics}
\begin{tabular}{ m{5em}  m{4cm} m{2cm} }
  \toprule
  Aspect & Metric & Results \\ 
      \cmidrule(lr){1-3}
  Safety & Success rate of path completion(\%)  & 90 \\ 
      \cmidrule(lr){1-3}
  Algorithm Efficiency  &  Flight Time (s) \newline Inter-step learning time (s) \newline inter-episode learning time (s) &
  11.1 \newline 6.61e-06 \newline 5.76e-05 \\ 
  \bottomrule
\end{tabular} \label{tab:h2_metrics}
\end{table}

\begin{figure}[hbt!]
    \centering
    \begin{tikzpicture}
    \begin{groupplot}[
    group style = {group size = 1 by 1, horizontal sep=1.5cm, vertical sep=1.5cm,},
    width = 1.0\columnwidth, 
    height = 4.0cm,
    xmin=0, 
    xmax=100, 
    ymin=0, 
    ymax=100,
grid=none, axis line style={draw=none}, 
    tick style={draw=none}, 
    clip=false,
    xticklabel=\empty, 
    yticklabel=\empty,
    axis on top=true,
    clip marker paths=true
    ]

    \nextgroupplot[]
        \node[opacity=1.0] at (axis cs:50,50) () {\includegraphics[width=.99\columnwidth, height=3cm]{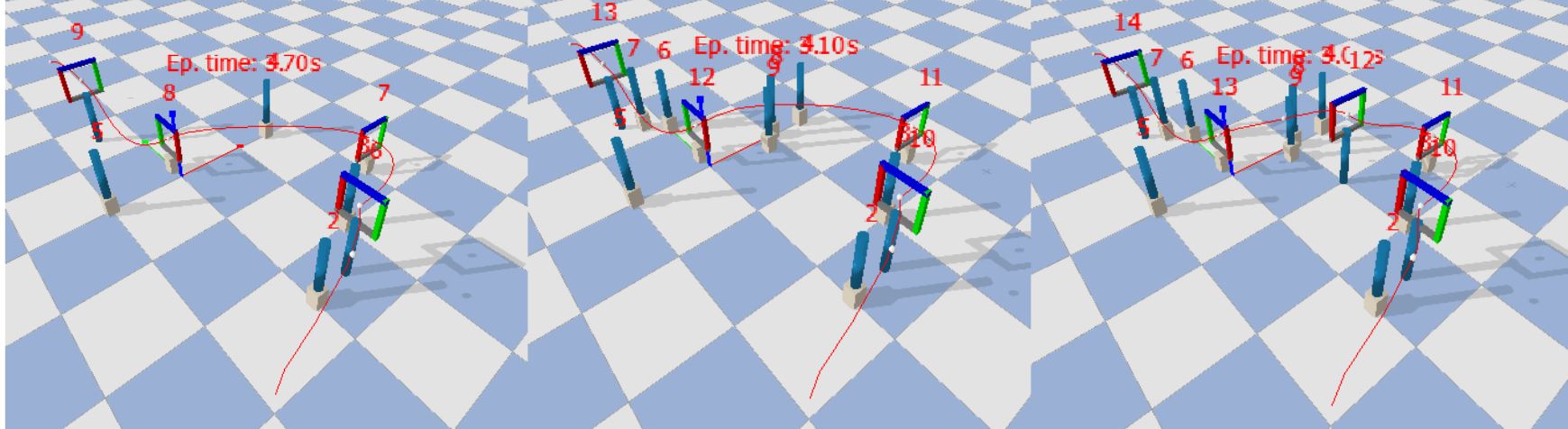}};
    \end{groupplot}

\end{tikzpicture}
\vspace{-1em}     \caption{Examples scenarios generated for the competition.}
    \label{fig:h2_test_cases}
\end{figure}

In addition to the metrics in Table~\ref{tab:h2_metrics}, there were other metrics, such as the density of the obstacles in the scenarios, that were used to test the robustness of the obstacle avoidance algorithm, which was done by creating multiple test cases which involved different locations and numbers of obstacles relative to the gate and checking their success rate. Figure~\ref{fig:h2_test_cases} shows an example of the test cases that were used to test the robustness of the solution.

After testing, Team H$^2$ also identified several potential areas for improving the solution. On the state estimation front, it would be useful to have Kalman filtering for the feedback of sensor data as well as full state feedback for the implementation of observations for the trajectory tracker. Clustering of obstacles could also be used to create larger radii for the drone to avoid, in denser environments. Reinforcement learning could also be explored to learn the full state commands needed for handling environmental disturbances, and local re-planning of trajectories could also be considered with sufficient onboard computing for online obstacle avoidance. 

In addition, the metrics used for this competition were averaged out over different scenarios, with varying numbers of obstacles and gates. With more time, it would be useful to examine the sensitivity of these metrics to the number of obstacles and gates in order to measure the scalability of the solution on larger-scale problems.

\section{Competition and Sim2real Results}
\label{sec:sim2real-results}

\begin{figure*}[hbt!]
    \centering
\includegraphics[]{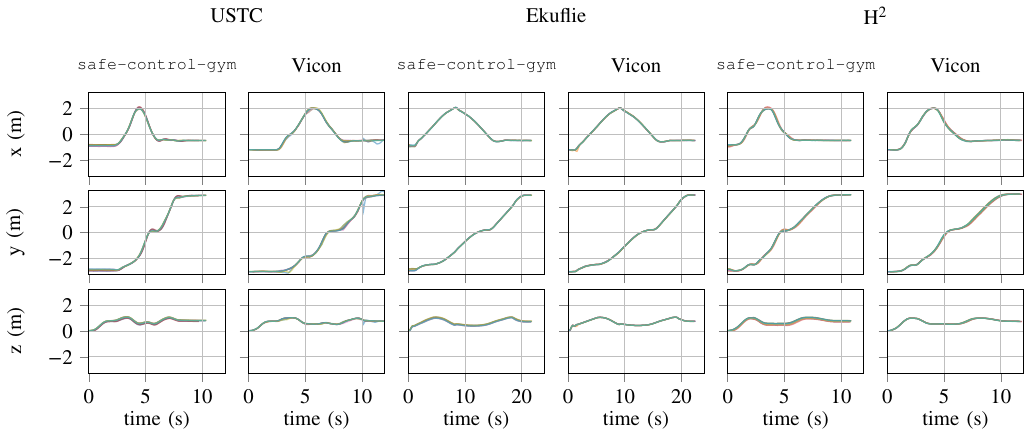}
    \caption{
    Comparison over 10 repeated simulations ({\small \texttt{safe-control-gym}}) and 10 real-life experiments (Vicon) of the trajectories followed by the Crazyflie commanded by the solutions of each of the 3 finalist teams (USTC, 1st and 2nd columns, Ekuflie, 3rd and 4th, H$^2$, 5th and 6th).
    }
    \label{fig:sim2real-res}
\end{figure*}

We performed simulations on a Lenovo P52 workstation laptop, equipped with 32GB of RAM, an Intel i7-8850H CPU, and an NVIDIA Quadro P2000 GPU, running 20.04.4 LTS. The same laptop was used, together with the University of Toronto's Vicon motion capture system, to run Crazyswarm~\cite{crazyswarm} for real-life experiments.

Figure~\ref{fig:sim2real-res} reports the positions, over time, in $x$, $y$, and $z$ of the Crazyflie drone in simulation (columns titled {\small \texttt{safe-control-gym}} and in the real world experiments (as tracked by Vicon) for the top 3 solutions by teams USTC, Ekuflie, and H$^2$.
We observe the pair-wise similarity, along the columns, on each row.
This shows that there is a close match between the trajectories followed by the simulated and the real-life quadrotor.
This includes both the directions traveled the most (i.e. $x$ and $y$, for $\sim$4m) as well as $z$ (it is important to observe that the gates were placed at different heights).
The results show an effective sim2real transfer for all solutions, especially w.r.t. their safety that, in our competition setup was expressed through multiple position constraints.

\begin{figure}[hbt!]
    \centering
\includegraphics[]{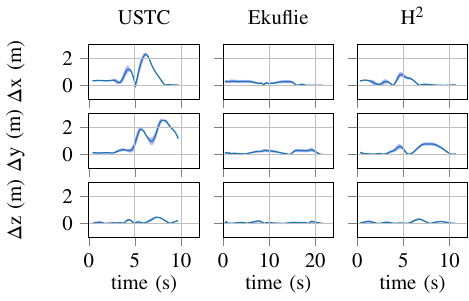}
    \caption{
    Average position error (over 10 repeated simulations and real-life experiments) resulting from the sim2real transfer performance of the competition's hardware/software infrastructure for the 3 finalist teams.
    }
    \label{fig:sim2real-res-diff}
\end{figure}

In Figure~\ref{fig:sim2real-res-diff}, we directly quantify the sim2real performance of each of the 3 solutions by plotting the average position error (again, in $x$, $y$, and $z$) between 10 repeated simulations and 10 repeated real-life experiments.
The lowest sim2real position RMSE ($0.21$, $0.19$, $0.07$) is registered by team Ekuflie, however, theirs is the slowest solution ($\sim$21s). Team H$^2$ achieved similar RMSE with a much faster ($\sim$11s) completion time.
Team USTC's solution was similarly quick but also resulted in a much higher sim2real mismatch: once transferred to the real system, the controller could not catch up with its simulated counterpart, lagging a few seconds behind.

\begin{table}
    \centering
    \caption{Scores (Safety \& Performance; Data Efficiency) and Final Rankings}
    
\begin{tabular}{ccccccc}
    \toprule
      & lvl0 & lvl1 & lvl2 & lvl3 & sim2real & Tot. (Rank) \\
    \cmidrule(lr){1-7}

    USTC & 20; 5 & 20; 5 & 5; 5 & 5; 5 & 10; 5 & 85 (3) \\
    Ekuflie & 10; 10 & 10; 10 & 20; 10 & 20; 10 & 5; 10 & 115 (2) \\
    H$^2$ & 5: 20 & 5; 20 & 10; 20 & 10; 20 & 20; 20 & 150 (1) \\

    \bottomrule
\end{tabular}
     \label{table:rank}
\end{table}

Table~\ref{table:rank} summarizes the point assignments that each of the top three finalists (Sections~\ref{sec:ustc}, \ref{sec:ekuflie}, and \ref{sec:h2}) received, according to the scoring system in Section~\ref{sec:score}. Winning team H$^2$ consistently outperformed its adversaries in terms of data efficiency and in the sim2real stage of the competition. Ekuflie's performance was especially good in the more complex (2 and 3) simulation levels but could not effectively transfer to the sim2real stage of the competition because of the lack of real-time constraints on the interstep optimization stage (that led to the controller falling back on a more conservative approach).
Team USTC had the best performance in the easier simulation levels but fell behind in the more complex ones.

 \section{Lessons Learned and Future Outlook}
\label{sec:lessons}

With respect to demographics, a survey run after the completion of the competition showed that 55.6\% of the participants came from Asia and 44.4\% from the Americas (we should note that the synchronous part of the competition---final evaluations and awards---was held in a timezone that was especially inconvenient to European and African participants).
The survey also showed that team sizes ranged from 1 to 7 people, they comprised 33.3\% of industry professionals, 44.4\% of master's, and 22.2\% of undergraduate students. One-third of the respondents to the survey stated that they worked on their solution for over 4 months (i.e., from the earliest public announcement of the competition codebase).

On a scale from 1 to 5, the participants scored the competition relevance as 4.5 for \emph{robotics}, 3.4 for \emph{safety}, and 3.1 for \emph{learning}. The lower scores given to the learning aspect were supported by the following comments:
``the scenarios were not diverse enough to make it a requirement and the scoring system would penalize solutions that spent [...] time learning to improve performance.''; ``I think that the plausibility for reinforcement learning-based methods [...] for the solution was relatively less as compared to the traditional control-based approaches [...]. Using adaptive control might suffice for tackling the external disturbances/rejections.''; ``Having a Mellinger controller in the loop restricts the [...] controller's ability to shape the system response and thus no attempt was made to learn and correct for uncertainty or bias in model dynamics.''

Overall, the difficulty level and the scoring systems were considered either fair or only slightly too difficult/convoluted by 88.8\% and 77.7\% of the survey respondents, respectively.
The entirety of the respondents stated that they would like to see the same competition re-proposed in future editions of IROS in its current format (66.7\%) or with only minor modifications (33.3\%).

We believe that the software infrastructure we built for the competition is also useful for the effective and efficient validation of many other aerial research topics. To this end, we aim to improve the infrastructure in the future further.
With regard to the sim2real component, we observed that the competition infrastructure measured (and raised warnings associated with) the run-time of the learning and optimization methods of each solution controller. However, it did not contain provisions for their clipping. This allowed for solutions' complexity (e.g., Ekuflie's) to remain potentially unbounded, resulting, on occasions, in different simulation and real-world performance.
In future iterations of the competition, we will implement strict run-time constraints in the simulation code to avoid this phenomenon.
Another ongoing plan is to integrate some of our developed components for wider use. Specifically, we plan to integrate our additions in \texttt{\small pycffirmware} to the official Python bindings and the physics simulation of \texttt{\small safe-control-gym} to Crazyswarm2.

\balance

\balance
\end{document}